\pdfoutput=1
\documentclass[11pt]{article}

\usepackage{ACL2023}

\usepackage{times}
\usepackage{latexsym}
\usepackage{multirow}
\usepackage{makecell}
\usepackage{graphicx}
\usepackage{enumitem}
\usepackage[graphicx]{realboxes}
\usepackage{listings}
\usepackage{xcolor}

\definecolor{codegreen}{rgb}{0,0.6,0}
\definecolor{codegray}{rgb}{0.5,0.5,0.5}
\definecolor{codepurple}{rgb}{0.58,0,0.82}
\definecolor{backcolour}{rgb}{0.95,0.95,0.92}

\lstdefinestyle{mystyle}{
    backgroundcolor=\color{backcolour},   
    commentstyle=\color{codegreen},
    keywordstyle=\color{blue},
    numberstyle=\tiny\color{codegray},
    stringstyle=\color{codepurple},
    basicstyle=\ttfamily\footnotesize,
    breakatwhitespace=false,         
    breaklines=true,                 
    captionpos=b,                    
    keepspaces=true,                 
    numbers=left,                    
    numbersep=5pt,                  
    showspaces=false,                
    showstringspaces=false,
    showtabs=false,                  
    tabsize=2
}

\usepackage{amsthm}
\newtheorem{theorem}{Theorem}
\newtheorem{remark}[theorem]{Finding}

\usepackage[T1]{fontenc}
\usepackage[utf8]{inputenc}

\usepackage{microtype}

\usepackage{inconsolata}
\usepackage{soul}

\title{A Static Evaluation of Code Completion by Large Language Models}

\author{Hantian Ding, Varun Kumar, Yuchen Tian, Zijian Wang, Rob Kwiatkowski, \\ 
{\bf Xiaopeng Li}, {\bf Murali Krishna Ramanathan}, {\bf Baishakhi Ray},\\ {\bf Parminder Bhatia}, 
{\bf Sudipta Sengupta}, {\bf Dan Roth}, {\bf Bing Xiang}\\
    AWS AI Labs \\
    \texttt{\{dhantian, kuvrun, tiayuche, zijwan, robkwiat, xiaopel} \\ 
    \texttt{mkraman, rabaisha, parmib, sudipta, drot, bxiang\}@amazon.com}
}

\begin{document}
\maketitle
\begin{abstract}
Large language models trained on code have shown great potential to increase productivity of software developers. Several execution-based benchmarks have been proposed to evaluate functional correctness of model-generated code on simple programming problems. Nevertheless, it is expensive to perform the same evaluation on complex real-world projects considering the execution cost. On the contrary, static analysis tools such as linters, which can detect errors without running the program, haven't been well explored for evaluating code generation models.
In this work, we propose a static evaluation framework to quantify static errors in Python code completions, by leveraging Abstract Syntax Trees.
Compared with execution-based evaluation, our method is not only more efficient, but also applicable to code in the wild. 
For experiments, we collect code context from open source repos to generate one million function bodies using public models.
Our static analysis reveals that Undefined Name and Unused Variable are the most common errors among others made by language models.
Through extensive studies, we also show the impact of sampling temperature, model size, and context on static errors in code completions.  

\end{abstract}

\section{Introduction}

Automatic code completion by large language models trained on numerous code repositories has demonstrated great potential in accelerating software development. Code assistant services powered by these models provide developers with code suggestions following the current context in real-time. However, it has been shown that about 70\% of the suggestions are discarded by users in a recent study \cite{SIGPLAN/copilot_assess}. Even worse, misleading recommendations can lead to failure in completing programming tasks \cite{CHI/copilot_study}.
Therefore, it is important to understand the weakness of current code generation models through comprehensive evaluation and analysis.

\begin{figure}[t]
    \centering
    \includegraphics[width=0.45\textwidth]{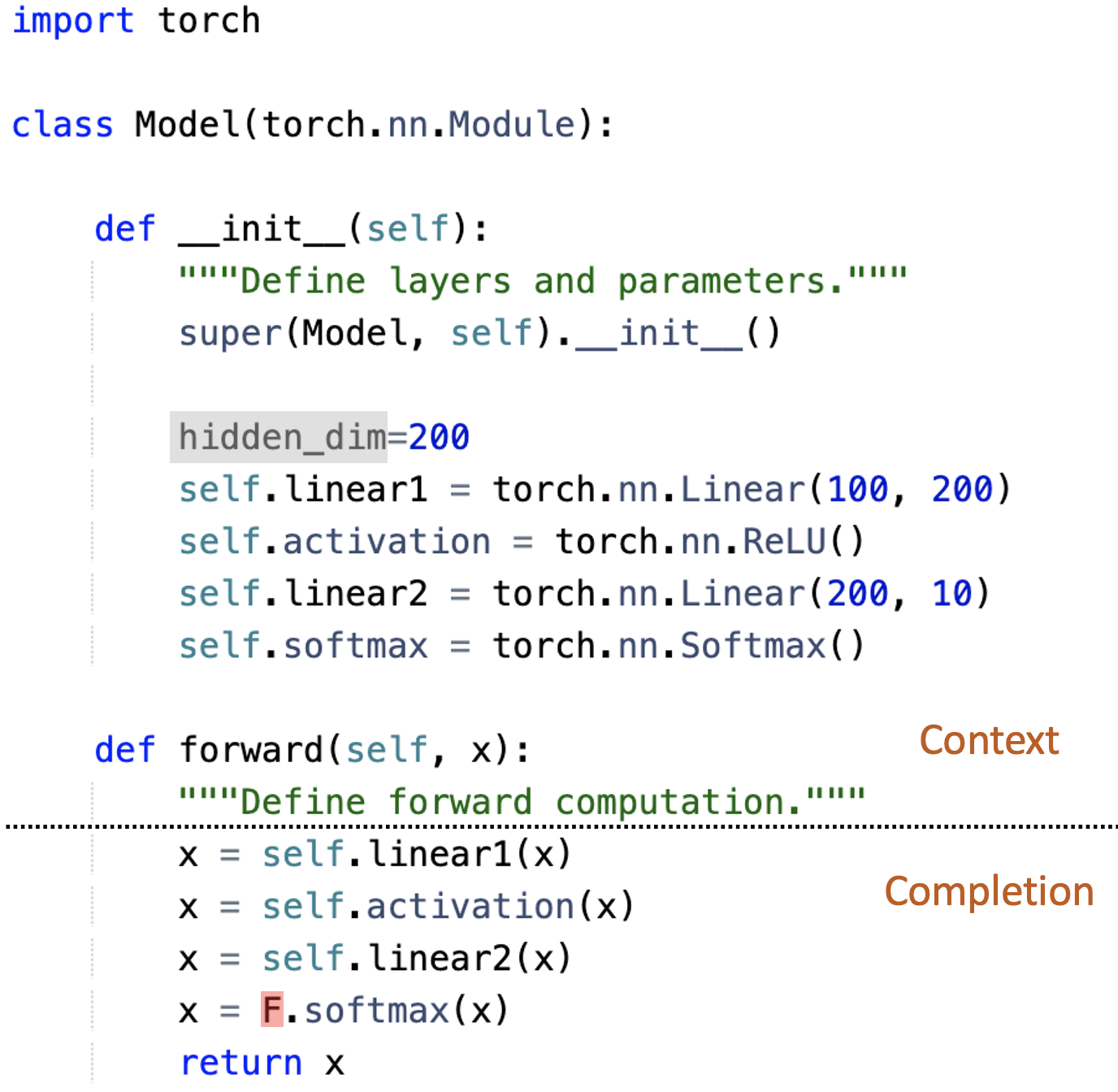}
    \caption{A function completion example, with an Unused Variable error (gray) in context, and an Undefined Name error (red) in completion.}
    \label{fig:function_completion}
\end{figure}

Recently, execution-based evaluation has become increasingly popular, where model-generated code is executed with unit tests to check functional correctness. Several benchmarks have been proposed along this direction, such as HumanEval \cite{codex}, MBPP \cite{mbpp}, MBXP \cite{mbxp}, CodeContests \cite{alphacode}, and DS-1000 \cite{ds-1000}. Although these benchmarks are highly reliable and accurate, they only focus on well-defined algorithmic and data science problems, which do not reflect the need in general software development. Running execution-based evaluation with real-world codebases is, however, prohibitively expensive because each project requires a different setup and the computation cost is potentially unbounded.

In contrast to the execution-based approach, \textit{static program analysis} (or \textit{static analysis}) can analyze programs without executing them. Although static analysis is usually unable to determine functional correctness, it covers a large collection of static error types, such as undefined names or unused variables that are illustrated in Figure \ref{fig:function_completion}. More importantly, the analysis can be very fast and does not require any project specific environment setup, which allows us to evaluate model completions for complex real-world code at large scale. Static analysis tools such as linters have been widely used, for example in code editors, to examine human-written code, but their value in evaluating code generation models has not been well explored yet.

In this work, we propose a static evaluation framework for Python language. Code snippets are first parsed into Abstract Syntax Trees (ASTs) and then analyzed by Pyflakes\footnote{\url{https://github.com/PyCQA/pyflakes}}, a popular static analysis tool for Python. To simulate real-world use cases of auto completion, we collect code from public Github repositories to build a function completion dataset of 100K problems. In each problem, we randomly mask out a function body in a Python file and ask the model to complete it given the preceding context up until the function header. We then evaluate public models by sampling 10 completions for each problem, resulting in one million generations for each model and sampling temperature, which will be examined by our static evaluation pipeline.

During AST parsing, we find most of the errors arise from incomplete generations that hit the max length limit. Otherwise, models of all sizes perform quite well in producing parsable codes. Moving forward, Pyflakes analysis reveals that Undefined Name and Unused Variable are the most prominent static errors in model-generated code. We also observe higher temperatures consistently lead to more errors. Scaling up the model, while able to reduce errors of many types, do not show a clear benefit for preventing undefined names. Through a more fine-grained classification, we find larger models generate fewer undefined variables but more undefined methods, which add up to a mixed result. Finally, we demonstrate that errors in context can lead to errors of the same type in generation, which is likely a consequence of large language models' in context learning capability.

In summary, our main contributions include the following. (1) We propose a static evaluation framework for code completion. (2) Our evaluation on public models reveals common static errors and how they are impacted by various factors such as temperature, model size, and context.  

\section{Background}

\textbf{Code Generation with Transformers}
Over recent years, it has become increasingly popular to train Transformer-based language models on source code \cite{feng-etal-2020-codebert, ahmad-etal-2021-unified, wang-etal-2021-codet5, codexglue, guo-etal-2022-unixcoder} to support software engineering tasks \cite{iyer-etal-2018-mapping, 10.1145/3340544}. In particular, several decoder-only transformer models have been developed to facilitate code generation, such as Codex \cite{codex}, CodeGen \cite{codegen}, Incoder \cite{incoder}, and AlphaCode \cite{alphacode}. These pretrained causal language models can be used to predict the continuation of input code without any finetuning. 

%\noindent
\textbf{Abstract Syntax Tree}
An Abstract Syntax Tree (a.k.a., AST) is used to represent a source code in a concise tree form. 
By discarding unnecessary details of the underlying code and its corresponding parsed tree, 
AST only presents the main structural content of the source code following the language grammar~\cite{aho2007compilers}. 

%\noindent
\textbf{Static Analysis} 
Static analysis is a common way to detect software bugs without executing the program~\cite{ayewah2008using, chess2004static, chess2007secure, zheng2006value}. Static analyzers tend to detect bugs by analyzing the static code text, its AST, documentation, etc. The users usually need to specify the error patterns and static analyzers use different AST, graph, and path analysis to find those patterns in the code. There are a plethora of static analysis tools and they can detect a wide range of errors depending on the specified patterns~\cite{emanuelsson2008comparative}. 
For example, Linter is a popular tool that checks for coding style errors and thus, tries to enforce a coding standard~\cite{van2021prevalence}.

\section{The Function Completion Dataset}\label{offline_data}

\begin{figure*}[t]
    \centering
    \includegraphics[width=0.9\textwidth]{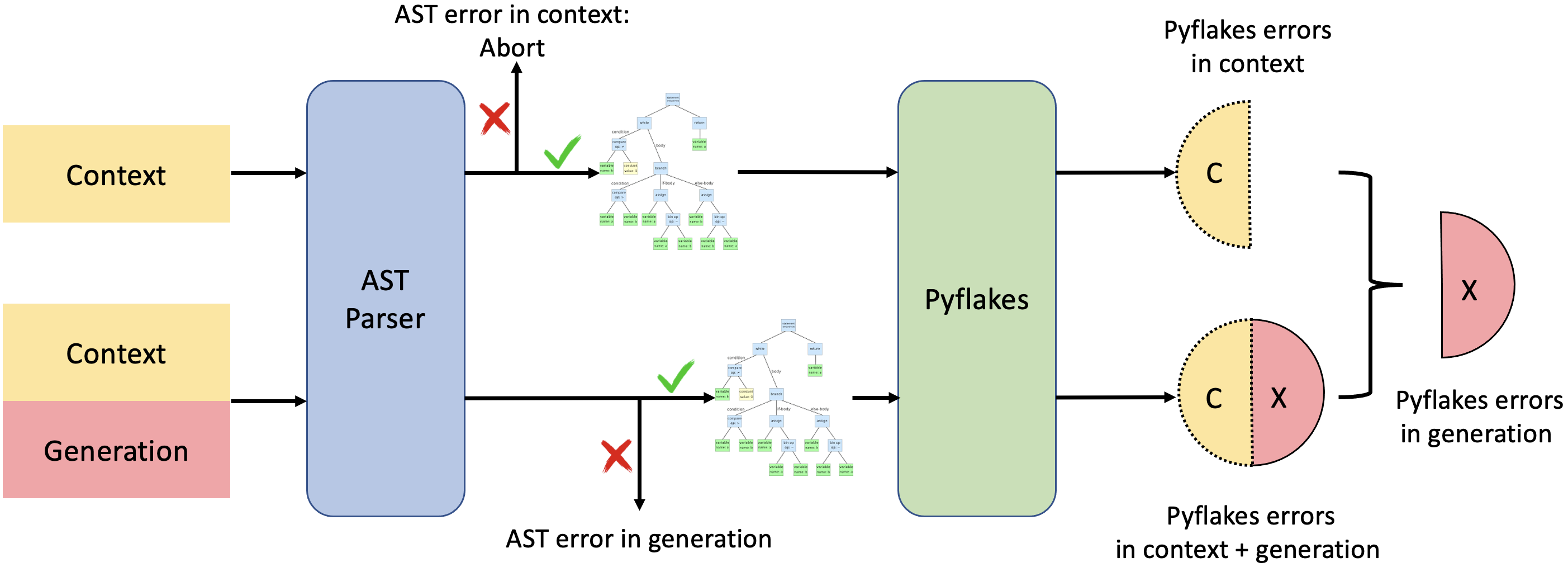}
    \vspace{-0.05in}
    \caption{Evaluation pipeline. \textbf{Left}: We parse [context] and [context + generation] into ASTs. If [context] is not parsable, we stop without reporting any error on generation. If [context] is parsable, but [context + generation] is not, we report the AST error in generation. \textbf{Right}: If both are parsable, we run Pyflakes on the trees, which reports errors in [context] and errors in [context + generation]. Taking the difference gives us errors in generation.}
    \vspace{-0.15in}
    \label{fig:method}
\end{figure*}

We introduce the \textit{function completion} task, which is one of the most important use cases of auto completion services.
Given an input code snippet that ends with a function signature plus an optional docstring, the model is asked to generate the function body. 
Previous works on code completion \cite{codexglue, DBLP:conf/kdd/SvyatkovskiyZFS19} have mainly focused on single-line completion. However, a single line is often too short to reveal models' capability in writing syntactically correct code. We believe function, as the fundamental building block in most programming languages, better serves this purpose. 

Software developers use code generation models as black-box services on a diverse set of coding projects. To better simulate the real-world scenario, we build an evaluation set by sampling from public Github repositories. Specifically we collected permissively licensed Python code in repositories that were created between April, 2022 and August, 2022. The selection criterion precludes any chronological overlap between our evaluation data and the training data of models to be tested in this work.\footnote{CodeGen models were trained on data up until Oct, 2021.} 

The collected Python codes are reformatted as function completion problems. We first use tree-sitter\footnote{\url{https://tree-sitter.github.io/tree-sitter/}} to parse the whole file to identify all the functions. Then a function that contains a docstring is randomly selected. The code from the beginning of the file up until the end of the docstring is used as the context, and the function body is considered as the groundtruth. The rest of the file is discarded. At test time, we prompt the model with the context part as input, and let the model generate the function body. We choose only functions with docstrings so that context is well-defined and model can generate meaningful code completions. We further select test samples whose context length is between 64 and 768 tokens, and groundtruth length is shorter than 256 tokens, to match our model generation setting. 
Our final evaluation set consists of 100K function completion problems.

\section{Static Error Analysis}

We propose an evaluation pipeline to detect errors in function completions generated by models, illustrated in Figure \ref{fig:method}. Suppose the model generates a completion $x$ given the input context $c$. We cannot directly analyze $x$ which is partial code without context. Meanwhile, $c$ may also contain errors especially in real-world cases. Therefore, we perform our analysis in two passes. We first check $c$ for any errors in the input that need to be excluded, and then do another pass on the full code $(c, x)$, the concatenation of the context and model completion. Any error that is identified in $(c, x)$ but not in $c$ must arise from $x$, or in other words, be generated by the model.
More specifically, we conduct the following two steps of analysis for Python code. 

\subsection{AST parsing}
In the first step, we parse both $c$ and $(c,x)$ into abstract syntax trees using Python's native \textit{ast} module. If the code is parsable, an AST will be returned. Otherwise, a syntax error is captured.
Based on the parsing outcomes, we take the following actions:

\begin{enumerate}[leftmargin=*]
\vspace{-0.05in}
\item If $c$ is not parsable, we are unable to conclude any error in generation. Empirically this rarely happens, as we will show in the next section.
\vspace{-0.05in}
\item If $c$ is parsable but $(c,x)$ is not, then we can confirm the reported syntax error is caused by model generation. However, notice that only one error will be returned even if there are multiple, due to the nature of AST parsing. 
\vspace{-0.05in}
\item If both $c$ and $(c,x)$ are parsable, there's no AST error in model generation. The ASTs will be used for static analysis in the next step.
\vspace{-0.05in}
\end{enumerate}

\subsection{Static analysis with Pyflakes}
If both $c$ and $(c,x)$ can be parsed into ASTs, we perform static analysis using Pyflakes. Pyflakes is a static analysis tool that checks a Python source file for errors by examining the AST. One advantage is that the analysis does not rely on dependencies of the source file, which is important given the diversity of packages used in real-world code. 
We run Pyflakes on $c$ and $(c,x)$ to identify errors in context and in full code. Errors that are detected in $(c,x)$ but not in $c$ are considered as introduced by model completion. 
\section{Experiments}\label{sec:offline}
With the proposed pipeline we conduct error analysis for CodeGen models \cite{codegen} on the test set described in Section \ref{offline_data}, and present the analysis results. 

\subsection{Experiment Setup}
We evaluate CodeGen-mono models of all sizes, ranging from 350M to 16B. We generate function completions using nucleus sampling with top-p 0.95. Sampling temperature is varied between 0.2 and 0.8 for the 2B model, and fixed to 0.4 for the rest models. We sample 10 generations for each problem, which results in one million code completions for each model and temperature. The maximum generation length is 256 tokens. Generated code completions are then passed to our static evaluation pipeline built with Python 3.8 and Pyflakes 3.0.1. Evaluating one million generations takes only a few hours on a single CPU thread, and can be fully parallelized for acceleration.

\subsection{Validation of Model Output}

While we mainly focus on static errors in this study, it is also important to validate that the models do generate relevant code. A counter-example would be to generate a single line of "return" for every function signature, which is syntactically correct but not meaningful at all. Towards this end, we calculate the edit similarity between model generation and groundtruth, and compare against Pass@1 from HumanEval \cite{codex} which is a popular execution-based benchmark to evaluate code generation models. Specifically, for both datasets we generate 10 samples per problem, and report the averaged edit similarity or pass rate over all generations. As shown in Table \ref{table: val}, models of all sizes and temperatures are able to achieve reasonable edit similarity on the function completion dataset, which means the generations are semantically relevant. Moreover, edit similarity and HumanEval Pass@1 both improve as the model scales up, highlighting that model scale is crucial for accurate code generation. Finally, the strong positive correlation between the last two columns shows that edit similarity on the function completion dataset can be used as an alternative metric for model comparison.

\begin{table}[t]
\centering
\begin{tabular}{lc|c@{\hskip 0.05in}c}
\hline
\multicolumn{1}{c}{Model}   & Temp                 & \thead{Edit\\Similarity} & \thead{HumanEval\\Pass@1} \\ \hline
CodeGen-16B                 & \multirow{4}{*}{0.4} & 72.07           & 31.83  \\
CodeGen-6B                  &                      & 68.76           & 26.46  \\
CodeGen-2B                  &                      & 64.83           & 23.72  \\
CodeGen-350M                &                      & 56.47           & 12.62  \\ \hline
\multirow{4}{*}{CodeGen-2B} & 0.2                  & 65.10           & 25.06  \\
                            & 0.4                  & 64.83           & 23.72  \\
                            & 0.6                  & 64.09           & 21.28  \\
                            & 0.8                  & 62.62           & 17.56  \\ \hline
\end{tabular}
\vspace{-0.05in}
\caption{Edit similarity on function completion dataset and Pass@1 on HumanEval, of CodeGen models across different sizes and temperatures. (1) Edit similarity and HumanEval Pass@1 are positively correlated across different settings, which justifies edit similarity can be used as an alternative metric for model evaluation. (2) As expected, larger models have better edit similarity (a proxy to accuracy) on function completion task.  }
\vspace{-0.15in}
\label{table: val}
\end{table}

\subsection{AST Results}

\begin{table*}[t]
\centering
\begin{tabular}{lc|cccccc}
\hline
\multicolumn{1}{c}{Model}   & Temp                 & Total   & EOF     & Non EOF & \thead{Invalid\\Syntax} & \thead{"print"\\Missing\\Parentheses} & \thead{Keyword\\Argument\\Repeated} \\ \hline
CodeGen-16B                 & \multirow{4}{*}{0.4} & 7.330\% & 7.236\% & 0.094\% & 0.042\%        & 0.041\%         & 0.004\%        \\
CodeGen-6B                  &                      & 7.446\% & 7.253\% & 0.193\% & 0.081\%        & 0.094\%         & 0.006\%        \\
CodeGen-2B                  &                      & 7.272\% & 7.177\% & 0.095\% & 0.052\%        & 0.018\%         & 0.008\%        \\
CodeGen-350M                &                      & 8.703\% & 8.593\% & 0.110\% & 0.041\%        & 0.016\%         & 0.028\%        \\ \hline
\multirow{4}{*}{CodeGen-2B} & 0.2                  & 8.067\% & 7.982\% & 0.085\% & 0.045\%        & 0.018\%         & 0.008\%        \\
                            & 0.4                  & 7.272\% & 7.177\% & 0.095\% & 0.052\%        & 0.018\%         & 0.008\%        \\
                            & 0.6                  & 6.823\% & 6.713\% & 0.110\% & 0.060\%        & 0.020\%         & 0.008\%        \\
                            & 0.8                  & 7.496\% & 7.337\% & 0.159\% & 0.085\%        & 0.029\%         & 0.014\%        \\ \hline
\end{tabular}
\vspace{-0.05in}
\caption{Percentages of AST errors across different model sizes and temperatures. We show (1) total AST errors; (2) errors at the end of file (EOF); (3) errors not at EOF; (4) top 3 non-EOF errors. Models generally perform well at AST level except for EOF errors caused by max generation length limit.}
\vspace{-0.15in}
\label{table: ast}
\end{table*}

We run AST parsing and find there are only 0.42\% cases with unparsable context that need to be discarded. For the rest, we report percentage of generations with AST errors in Table \ref{table: ast}. A full list of error types is included in Appendix \ref{sec:appendix_error_category}. For each type, we also show a code example in Appendix \ref{sec:appendix_example}. 

While there are about 7-8\% of unparsable generations, most of the parsing errors happen at the end of file (EOF), which means the generated code is incomplete due to the 256 max token limit. Extending generation length may help reduce EOF errors, but will require more computation and increase the perceived latency of the auto-completion service. 

On the other hand, non-EOF errors only account for a tiny fraction, usually around 0.1-0.2\%, which indicates CodeGen models can generally follow the abstract syntax grammar to produce parsable codes, regardless of model size and temperature. 

\begin{remark}
Codes generated by models, unless incomplete, are mostly parsable into ASTs, regardless of model size or temperature.  
\end{remark}
We also show the top-3 non-EOF error types ranked by frequency, which are \textbf{Invalid syntax}, \textbf{Print Missing Parentheses}, and \textbf{Keyword Argument Repeated}. Notably, the first two categories are often related to Python's interpreter version. To illustrate, Python2-style print like \textit{print "abc"} will lead to Print Missing Parentheses in Python3. Another example is that using \textit{async} as a variable name will cause Invalid Syntax because \textit{async} has become a reserved word since Python3.7. Models learn to make such errors from their training data which consists of code written for different Python versions. In many cases, it is difficult for a model to infer the intended interpreter version directly from the limited context. 
An interesting future direction is to guide models to generate version-compatible code given the target environment. 
\begin{remark}
Interpreter version mismatch is one of the major reasons for non-EOF AST errors.
\end{remark}

\subsection{Pyflakes Results}

\begin{table*}[t]
\centering
\begin{tabular}{lc|cccccc}
\hline
\multicolumn{1}{c}{Model}   & Temp                 & \thead{Undefined \\ Name} & \thead{Unused \\ Variable} & \thead{FString\\Missing\\Placeholders} & \thead{Unused\\Import} & \thead{Redefined\\While\\Unused} & \thead{Undefined\\Local} \\ \hline
CodeGen-16B                 & \multirow{4}{*}{0.4} & 4.323\%       & 1.729\%        & 0.135\%                    & 0.107\%      & 0.131\%              & 0.047\%        \\
CodeGen-6B                  &                      & 4.374\%       & 1.775\%        & 0.089\%                    & 0.149\%      & 0.126\%              & 0.055\%        \\
CodeGen-2B                  &                      & 4.364\%       & 1.810\%        & 0.147\%                    & 0.150\%      & 0.146\%              & 0.065\%        \\
CodeGen-350M                &                      & 4.472\%       & 2.032\%        & 0.151\%                    & 0.173\%      & 0.155\%              & 0.095\%        \\ \hline
\multirow{4}{*}{CodeGen-2B} & 0.2                  & 4.206\%       & 1.751\%        & 0.125\%                    & 0.139\%      & 0.139\%              & 0.067\%        \\
                            & 0.4                  & 4.364\%       & 1.810\%        & 0.147\%                    & 0.150\%      & 0.146\%              & 0.065\%        \\
                            & 0.6                  & 4.711\%       & 2.000\%        & 0.188\%                    & 0.170\%      & 0.159\%              & 0.076\%        \\
                            & 0.8                  & 5.377\%       & 2.490\%        & 0.240\%                    & 0.247\%      & 0.184\%              & 0.086\%        \\ \hline
\end{tabular}
\vspace{-0.05in}
\caption{Percentages of Pyflakes errors across different model sizes and temperatures. Higher temperatures always lead to more errors in every category. On the other hand, larger models do not necessarily generate fewer errors.} %, and the trend varies by error type.}
\vspace{-0.15in}
\label{table: pyflakes}
\end{table*}

\begin{figure}[t]
    \centering
    \includegraphics[width=0.45\textwidth]{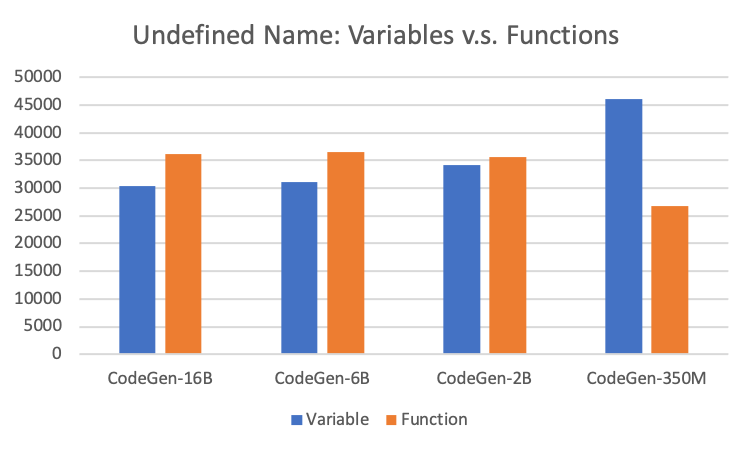}
    \vspace{-0.05in}
    \caption{Number of undefined variables versus undefined functions. Larger models generate more undefined functions but fewer undefined variables.}
    \vspace{-0.15in}
    \label{fig:undefined_names}
\end{figure}

We present frequencies of top 6 linter errors from Pyflakes in Table \ref{table: pyflakes}, with code examples in Appendix \ref{sec:appendix_example}. While Pyflakes also finds other problems in code, most of them are very sparse and thus less important, which we leave to Appendix \ref{sec:appendix_error_category}. Notice that one code snippet may contain multiple errors. We count each type only once in every test sample. 

Among all errors, \textbf{Undefined Name} and \textbf{Unused Variable} are the most common ones, where the model either calls a variable that is not defined, or defines a variable but never uses it. Closely related are \textbf{Unused Import}, \textbf{Redefined While Unused} and \textbf{Undefined Local}, which can be considered as special cases of the first two. Models also sometimes unnecessarily use f-strings by not giving any placeholder. It is worth pointing out that not all Pyflakes errors will impact execution. In fact among the six types, only Undefined Name and Undefined Local may cause runtime problems. However, all these errors can harm readability and maintenance which are critical for software development. Hence, it is important to address them to improve the quality of auto code completion.

Across sampling temperatures, we observe in every column that more errors are generated under higher temperatures, which is expected because generations in such cases are less confident. 
\begin{remark}
Higher temperature always leads to more errors of every type. 
\end{remark}

The impact of model size on error rate is less consistent though. For Unused Variable, Unused Import, and Undefined Local, error rate does decrease as the model scales up. However, the other three categories do not manifest such correlation. We investigate the underlying reason for this mixed result particularly in the case of Undefined Name. Notice that if an undefined name is a \textit{function call}, it can potentially be defined afterwards outside the current function completion scope. While not guaranteed, the model might be able to fix this error by itself if we allow generating longer code instead of only one function. In contrast, using a \textit{variable} without first defining it is usually a mistake. Even in some rare cases where the variable definition is made up correctly after the usage, such ordering is often less preferred in terms of coding style. In Figure \ref{fig:undefined_names}, we break down the undefined names into \textit{variables} and \textit{functions}. We find that larger models yield fewer undefined variables, but more undefined functions, which demonstrates that the correlation between error count and model size varies for different errors types. 
\begin{remark}
While larger models are more accurate code generators ~\cite{codegen}, scaling up model size does not lead to reduction in error counts for all error categories.
\end{remark}

\subsection{Correlation with Errors in Context}

We further study the correlation between errors in context and in generation. Denote by $c$ the input context, $x$ the model generation, $e$ the error type. We write $e\in c$ to mean $c$ contains an error of type $e$. For every $e$,\footnote{We omit Unused Import from Table \ref{table: pyflakes} because it is valid to have unused imports in the context that is yet to be completed.} we calculate $\textup{P}(e\in x|e\in c)$, the generation error rate when context contains the same type of error(s). We also report the relative ratio $\frac{\textup{P}(e\in x|e\in c)}{\textup{P}(e\in x|e\notin c)}$ to measure the impact of context. From Table \ref{table: context}, if the model observes errors in context, it is more likely to produce the same type of errors in generation, and the error rate can be amplified by 7$\sim$200 times depending on the type. This is possibly an undesired consequence of the \textit{in-context learning} capability of large language models. 

We also calculate $\textup{P}(e\in c|e\in x)$ to show how many of the generation errors co-occur with context errors. As indicated by the last column of Table \ref{table: context}, even though context errors can significantly amplify generations errors, the co-occurrences of two do not account for a large fraction. This implies problematic context is not the only factor for problematic generation, and it is often the case for models to produce errors even with correct context. 
\begin{remark}
Errors in context generally lead to more errors in generation.
\end{remark}

\begin{table}[t]
\centering
\small
\begin{tabular}{c@{\hskip 0.05in}c@{\hskip 0.05in}c@{\hskip 0.05in}c}
\hline
\thead{Error type}                   & \thead{$ {\scriptstyle \textup{P}(e\in x|e\in c)}$} & \thead{$\frac{\textup{P}(e\in x|e\in c)}{\textup{P}(e\in x|e\notin c)}$} & \thead{$ {\scriptstyle\textup{P}(e\in c|e\in x)}$} \\ \hline
\thead{Undefined Name}               & 26.33\%                                    & 7.80     & 25.99\%  \\
\thead{Unused Variable}              & 14.13\%                                    & 8.45     & 8.56\%   \\
\thead{FString Missing\\Placeholders} & 20.63\%                                    & 215.50   & 35.08\%  \\
\thead{Redefined\\While Unused}       & 2.44\%                                     & 21.16    & 22.30\%  \\
\thead{Undefined Local}              & 7.00\%                                     & 108.68   & 1.08\%  \\ \hline
\end{tabular}
\vspace{-0.05in}
\caption{Correlation between errors in context and in generation for the 2B model. First two columns indicate errors in context can amplify errors in generation; the last column shows not all generations errors can be attributed to context. Other models have similar results.
}
\vspace{-0.15in}
\label{table: context}
\end{table}
\section{Discussion}
We present a static evaluation framework for code completions generated by large language models. 
By utilizing the proposed framework, we conduct error analysis of CodeGen models on a large scale real-world Python evaluation set. 
Our experiment reveals common static errors made by pretrained models, as well as their frequency trend across model sizes and sampling temperatures.
By pointing out weaknesses of existing models, we hope our study also sheds light on future directions towards more accurate code generation.

There are a few limitations of this study. First, we focus on left-to-right code generation without considering right-side and cross-file context, which can be used to determine broader categories of errors with improved precision. Second, each static analysis tool has its own limitations. Thus, the presented analysis is limited by Pyflakes's accuracy and coverage to detect certain code issues.

\newpage
\bibliography{anthology,custom}
\bibliographystyle{acl_natbib}

\appendix
\newpage
 
~\newpage

\section{Full Error Categories}
\label{sec:appendix_error_category}
In addition to those discussed in Section \ref{sec:offline}, we list all error categories that can be detected in model generated code in our experiments, with a minimal frequency of 0.001\% by any of the models (i.e. 10 observations out of the total 1 million generations). 
%We additionally attach the absolute count of erroneous generations (out of 1 million) after the error name, for CodeGen-2B model under temperature 0.4. 

\vspace{0.2in}
\textbf{AST errors (EOF errors indicated by asterisk): }
\begin{enumerate}
    \item *unexpected EOF while parsing %(41913)
    \item *EOL while scanning string literal %(17952)
    \item *invalid syntax at EOF %(10284)
    \item *EOF while scanning triple-quoted string literal %(1516)
    \item invalid syntax not at EOF %(518)
    \item missing parentheses in call to "print" %(184) 
    \item keyword argument repeated %(82)
    \item leading zeros in decimal integer literals are not permitted; use an o prefix for octal integers %(49)
    \item unmatched ")" %(5)
    \item cannot assign to function call %(18)
    \item positional argument follows keyword argument %(6)
    \item expression cannot contain assignment %(4)
\end{enumerate}

\vspace{0.2in}
\textbf{Pyflakes issues: }
\begin{enumerate}
    \item undefined name %(43638)
    \item unused variable %(18099)
    \item f-string missing placeholder %(1471)
    \item unused import %(1495)
    \item redefined while unused %(1462)
    \item indentation error %(128)
    \item import shadowed by loop var %(146)
    \item raise not implemented %(125)
    \item invalid print syntax %(119)
    \item is literal %(40)
    \item string dot format extra positional argument %(25)
    \item multi value repeated key literal %(59)
    \item percent format positional count mismatch %(23) 
    \item tab error %(12)
    \item string dot format extra named arguments %(6)
    \item import star not permitted %(12)
    \item percent format unsupported format character %(6)
    \item assert tuple %(4)
    \item percent format extra named arguments %(10)
\end{enumerate}

\section{Examples for Top Error Types}
\label{sec:appendix_example}
Below we list one code example for each of the error categories shown in Table \ref{table: ast} and \ref{table: pyflakes}. Following the definition of function completion task, in every example, context is from the beginning until the end of the docstring of the last function, and model completion is the body of the last function. 

\centering
\begin{lstlisting}[language=Python,float=*,xleftmargin=0.5cm,caption=unexpected EOF while parsing (line 31)]
"""Secondary Structure dataset."""

import numpy as np
from megatron import print_rank_0
from .data import ProteinPredictionAbstractDataset
from .data import build_tokens_paddings_from_text

class SecondaryStructureDataset(ProteinPredictionAbstractDataset):
    def __init__(self,
                name: str,
                datapaths,
                tokenizer,
                max_seq_length: int):
        super().__init__('secondary_structure', name, datapaths, tokenizer, max_seq_length)


    def build_samples(self, ids, paddings, label, unique_id, seq_len):
        """Convert to numpy and return a sample consumed by the batch producer."""

        # Seperate inputs and labels in lists
        input_tokens = [self.tokenizer.tokenize(seq) for seq in ids]
        input_tokens = [token for seq in input_tokens for token in seq]
        labels = [self.tokenizer.tokenize(seq) for seq in label]
        labels = [label for seq in labels for label in seq]

        # Add special tokens
        input_tokens, labels = self.add_special_tokens(input_tokens, labels)

        # Truncate and pad
        input_tokens, labels, paddings = self.truncate_and_pad(input_tokens, labels,
                                                               self
\end{lstlisting}

\begin{lstlisting}[language=Python,float=*,xleftmargin=0.5cm,caption=invalid syntax (line 23)]
from pathlib import Path
from wasabi import msg
from .remote_storage import RemoteStorage
from .remote_storage import get_content_hash, get_command_hash
from .._util import load_project_config
from .._util import project_cli, Arg, logger


@project_cli.command("push")
def project_push_cli(
    # fmt: off
    remote: str = Arg("default", help="Name or path of remote storage"),
    project_dir: Path = Arg(Path.cwd(), help="Location of project directory. Defaults to current working directory.", exists=True, file_okay=False),
    # fmt: on
):
    """Persist outputs to a remote storage. You can alias remotes in your
    project.yml by mapping them to storage paths. A storage can be anything that
    the smart-open library can upload to, e.g. AWS, Google Cloud Storage, SSH,
    local directories etc.

    DOCS: https://xxx
    """
    for nlp in load_project_config(project_dir, {"directories": [ANIMAL_TRAIN_DIR]}:
    remote_storage = RemoteStorage.get(remote)
    for command in ["train"]:
        logger.debug(f"Uploading {command} to remote storage '{remote_storage}'")
        path = Path(project_dir) / ANIMAL_TRAIN_DIR / command
        upload_project(remote_storage, path)
\end{lstlisting}

\begin{lstlisting}[language=Python,float=*,xleftmargin=0.5cm,caption=Missing parentheses in call to "print" (line 6)]
def prod(F, E):
    """Check that the factorization of P-1 is correct. F is the list of
       factors of P-1, E lists the number of occurrences of each factor."""
    M = prod_of_prime_factors(F, E)
    if not all(i == 1 for i in M):
        print "Error in prod"
        print F, E
        return
    P = product(F)
    P_1 = 1
    for i in range(len(F)):
        P_1 *= F[i]**E[i]
    if P != P_1:
        print "Error in prod"
        print F, E
        print P
        print P_1
        return
\end{lstlisting}

\begin{lstlisting}[language=Python,float=*,xleftmargin=0.5cm,caption=keyword argument repeated (line 15)]
import unittest
from datetime import datetime, timezone

from dateutil.relativedelta import relativedelta

from bot.utils import time


class TimeTests(unittest.TestCase):
    """Test helper functions in bot.utils.time."""

    def test_humanize_delta_handle_unknown_units(self):
        """humanize_delta should be able to handle unknown units, and will not abort."""
        self.assertEqual(
            time.humanize_delta(datetime.utcnow(), datetime.utcnow() - relativedelta(months=1, months=2)),
            "1 month and 2 months"
        )
\end{lstlisting}

\begin{lstlisting}[language=Python,float=*,xleftmargin=0.5cm,caption=undefined name "factorial" (line 18)]
"""
This program will continually ask our user to give a number
and will calculate the factorial result of the number and print it on the console.

The program ends when the user enter the EXIT number.
"""

EXIT = -100


def main():
	"""
	This program will calculate the factorial result according to the number an user
	inputs.
	"""
	print('<<< Welcome to the Factorial Calculator! >>>')
	num = int(input('Enter a number: '))
	print('The factorial of {} is {}.'.format(num, factorial(num)))
	if num == EXIT:
		print('\n<<< Thank you for using the Factorial Calculator. >>>')
	else:
		main()
\end{lstlisting}

\begin{lstlisting}[language=Python,float=*,xleftmargin=0.5cm,caption=local variable "encoding\_check" is assigned to but never used (line 15)]
def check(full_path, encoding):
    assert type(full_path) == str, f'\'full_path\' is of {type(full_path)}. Only type \'str\' is acceptable.'
    assert full_path != "", "\'full_path\' is empty."
    assert type(encoding) == str, f'\'full_path\' is of {type(encoding)}. Only type \'str\' is acceptable.'
    assert encoding != "", "\'encoding\' is empty."

def file_read(full_path: str, encoding = "utf8"):
    '''
    Author: xxx

        Reads file at "full_path" and returns its data in a list.
    '''

    check(full_path, encoding)
    encoding_check = encoding
    full_path = full_path.strip()
    f = open(full_path, "r", encoding = encoding)
    lines = f.readlines()
    f.close()
    lines = [line.replace("\n", "") for line in lines]
    return lines
\end{lstlisting}

\begin{lstlisting}[language=Python,float=*,xleftmargin=0.5cm,caption=f-string is missing placeholders (line 15)]
import os
import json

from convinse.library.utils import store_json_with_mkdir, get_logger


class HeterogeneousAnswering:
    def __init__(self, config):
        """Initialize HA module."""
        self.config = config
        self.logger = get_logger(__name__, config)

    def train(self, sources=["kb", "text", "table", "info"]):
        """ Method used in case no training required for HA phase. """
        self.logger.info(f"No need to train.")
        pass
\end{lstlisting}

\begin{lstlisting}[language=Python,float=*,xleftmargin=0.5cm,caption="urllib.parse" imported but unused (line 17)]
import os
import urllib.parse
import sqlite3

SQL = """
SELECT p.ZAUTHOR, p.ZTITLE, e.ZTITLE, e.ZASSETURL, e.ZPUBDATE
from ZMTEPISODE e
join ZMTPODCAST p
    on e.ZPODCASTUUID = p.ZUUID
where ZASSETURL NOTNULL;
"""


def check_imports():
    ''' Prompts for password to install dependencies, if needed '''
    import os, importlib, importlib.util
    import urllib.parse

    # Check for dependency installs
    # Can be done more simply, but this way I can avoid importing anything from zmodel,
    # which is nice since I can see what's going on.
    for k, v in DEPS.items():
        try:
            importlib.import_module(k)
        except ImportError as e:
            importlib.util.find_spec(k)
            if importlib.util.find_spec(k) is None:
                os.system(f'pip install {v}')
\end{lstlisting}

\begin{lstlisting}[language=Python,float=*,xleftmargin=0.5cm,caption=redefinition of unused "dsl" from line 2 (line 6)]
import kfp.deprecated as kfp
from kfp.deprecated import components, dsl, compiler

def get_run_info(run_id: str):
    """Example of getting run info for current pipeline run."""
    import kfp.dsl as dsl
    client = kfp.Client()
    run = client.run_details(run_id)
    print(f"Run details:\n{run}")
    print(f"Pipeline details:\n{run.pipeline_runtime}")
\end{lstlisting}

\begin{lstlisting}[language=Python,float=*,xleftmargin=0.5cm,caption=local variable "cnt" defined in enclosing scope on line 16 referenced before assignment (line 18)]
"""Check for nonlocal and used-before-assignment"""
# pylint: disable=missing-docstring, unused-variable, no-init, too-few-public-methods

__revision__ = 0

def test_ok():
    """ uses nonlocal """
    cnt = 1
    def wrap():
        nonlocal cnt
        cnt = cnt + 1
    wrap()

def test_fail():
    """ doesn't use nonlocal """
    cnt = 1
    def wrap():
        cnt = cnt + 1 # [used-before-assignment]
    wrap()
\end{lstlisting}

\end{document}